\documentclass[letterpaper, 10 pt, conference]{ieeeconf}  

\IEEEoverridecommandlockouts                              

\usepackage{graphicx}
\usepackage{amsmath}
\usepackage{booktabs}
\usepackage{algorithm}
\usepackage{algorithmic}
\usepackage{cite}
\usepackage{amsfonts}
\usepackage{xcolor}
\usepackage{comment}
\usepackage[switch]{lineno}  %

\usepackage{caption, subcaption}
\newcommand{\fangjin}[1]{{\textcolor{red}{Fangjin: #1}}}

\title{\LARGE \bf
Large Scale Autonomous Driving Scenarios Clustering \\ with Self-supervised Feature Extraction
}

\author{Jinxin Zhao${^\star}$$^{1}$, Jin Fang${^\star}$$^{1}$, Zhixian Ye$^{1}$, and Liangjun Zhang$^{1}$

\thanks{$^{1}$ Baidu Research and National Engineering Laboratory of Deep Learning Technology and Application, China. For Jinxin Zhao, this work was done while he was at Baidu Research.}

\thanks{$^\star$ Joint first author.}

}

\begin{document}

\maketitle
\thispagestyle{empty}
\pagestyle{empty}

\begin{abstract}

The clustering of autonomous driving scenario data can substantially benefit the autonomous driving validation and simulation systems by improving the simulation tests' completeness and fidelity. This article proposes a comprehensive data clustering framework for a large set of vehicle driving data. Existing algorithms utilize handcrafted features whose quality relies on the judgments of human experts. Additionally, the related feature compression methods are not scalable for a large data-set. Our approach thoroughly considers the traffic elements, including both in-traffic agent objects and map information.
Meanwhile, we proposed a self-supervised deep learning approach for spatial and temporal feature extraction to avoid biased data representation. With the newly designed driving data clustering evaluation metrics based on data-augmentation, the accuracy assessment does not require a human-labeled data-set, which is subject to human bias. Via such unprejudiced evaluation metrics, we have shown our approach surpasses the existing methods that rely on handcrafted feature extractions.

\end{abstract}

\section{Introduction}

Autonomous driving technologies have been extensively studied and developed for the past decade, and the deployment of self-driving vehicles has gradually started \cite{waymolaunch}.
Vehicle driving data plays a critical role in accelerating the autonomous driving technology development
\cite{sun2020scalability}.
Essential autonomous driving components, especially simulation systems, use real-world vehicle driving data to generate close to real-world test cases to fully evaluate the autonomous driving software before deploying on the hardware platform
\cite{li2019aads,Koopman16}.  

A typical simulation generating and testing pipeline includes the following parts. First, real-world vehicle driving data, including the ego vehicle, other in-traffic agents, and the map information, is recorded. Second, the scenarios or types of driving situation is identified. Lastly, simulation tests are generated by reproducing the scenarios based on the recorded data. In such a process, providing a comprehensive list of scenarios for testing is critical.
Hence, an efficient clustering framework 
that can group vehicle driving data into different scenario types
plays a significant role in guaranteeing the effectiveness of autonomous driving simulation tests \cite{hauer2020clustering}. 
The benefits brought by an effective scenario clustering pipeline to the simulation system are as follows \cite{bach2016model,hauer2020clustering,hauer2019did}:
\begin{itemize}
    \item High-Fidelity: Such clustering system helps align the distribution of simulation test cases with the distribution of scenarios in the real world, ensuring the similarity between simulation and reality;
    \item Completeness: An efficient clustering algorithm can identify the simulation test cases that do not happen in the real world and those that could happen in the real world, even rarely (long-tail situations), guaranteeing the simulation test cases cover sufficient scenarios that happen in the real world;
    \item Efficiency: Such clustering framework can help remove redundant simulation test cases to avoid unnecessary simulation computation.
\end{itemize}

In addition to the autonomous driving simulation system, the self-driving planning module can as well benefit from a clustering algorithm, given it can be trained offline and can inference online. For example, the planning module of Baidu Apollo open-source software platform \cite{apolloplatform} utilizes a scenario-based architecture and selects a planning algorithm according to the scenario. Thus, concluding all the real-world scenario types from collected driving data and dynamically detecting scenario type on the road should be of great significance for such a planning system. 

By definition, the clustering algorithm solves an unsupervised machine learning problem, where similar data instances need to be grouped together \cite{xu2015comprehensive}. 
A clustering algorithm normally includes feature extraction, clustering design, and result evaluation \cite{xu2015comprehensive}. 
The feature extraction is one of the most critical processes for a clustering algorithm \cite{xu2005survey,nguyen2019feature}, which finds expressive representations for the data instances. As for the clustering of vehicle driving data, human selected features such as vehicle speed and distance between vehicles are normally used \cite{nguyen2019feature,bach2016model,kruber2018unsupervised}. However, 
such handcrafted features would introduce human mental bias
\cite{hauer2020clustering}.
These biased features would eventually result in deteriorated clustering results.
Similarly, the evaluation of clustering algorithms is a non-trivial task \cite{everitt2011cluster}. The internal evaluation takes advantage of the extracted features, thus affected by the feature extraction process. The external evaluation usually requires ground-truths cluster labels of the data instances \cite{amigo2009comparison}, where the ground-truths are generally unavailable. 

In this article, we introduce a framework of mapping the original data sequence into a feature representation. When an unsupervised clustering method is applied to the feature representation, similar data sequences will be grouped together. 

\begin{figure*}
    \centering
    \includegraphics[width=0.90\textwidth]{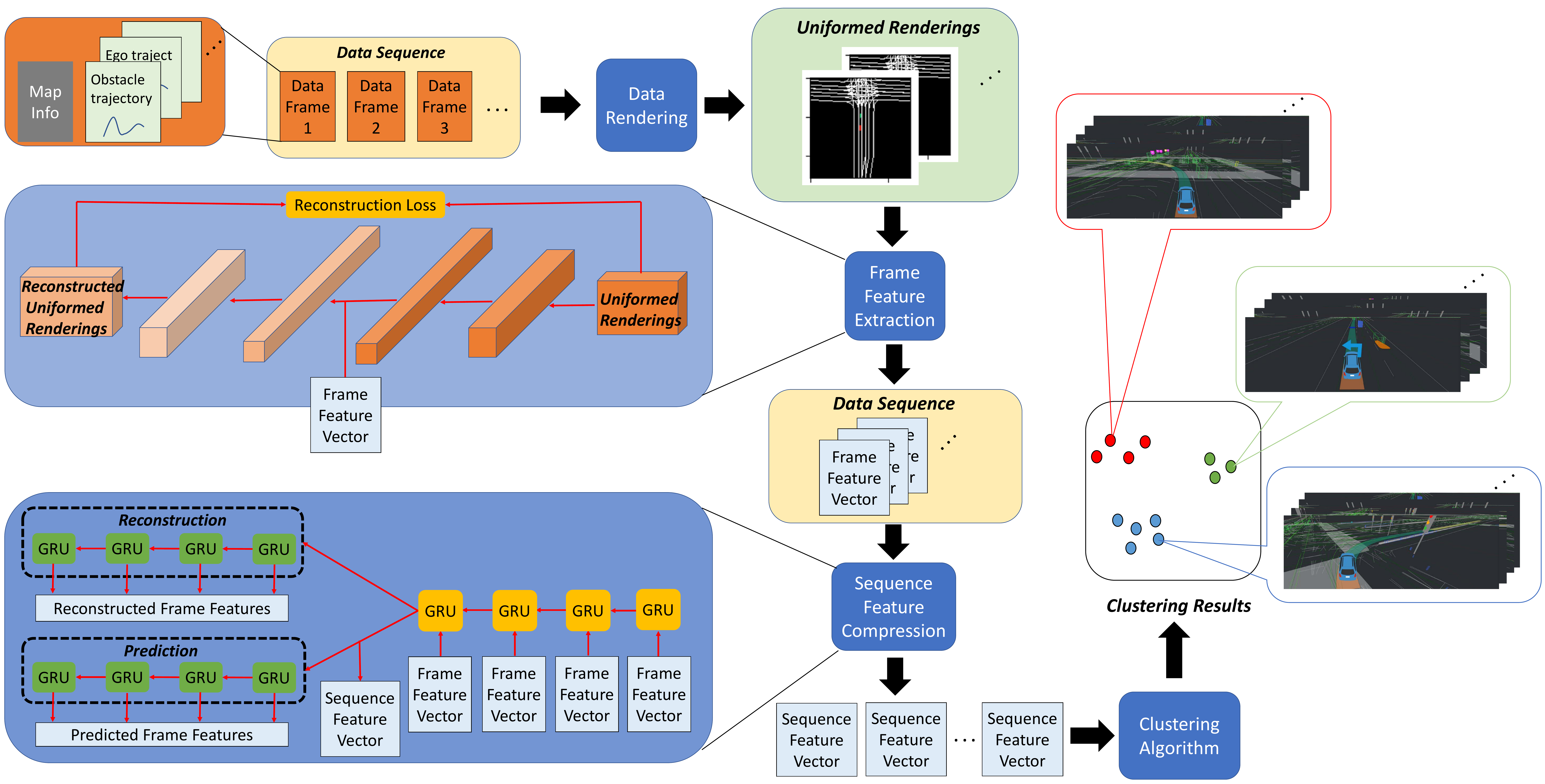} 
    \caption{The pipeline of vehicle driving data clustering. The information of both vehicle trajectories and map is first combined and unified through renderings. The rendering of each frame is first compressed for compact representation through CNN-VAE model as frame feature vector. 
    A sequence compression with RNN model further compresses a sequence of frame feature vectors into a sequence feature vector. A clustering algorithm then groups such compact sequence feature representations. The clustered sequences are visualized by simulation platform of Apollo \protect \cite{apolloplatform}.}
    \label{fig:framework}
\end{figure*}


The recorded vehicle driving raw data is typically composed of both the ego vehicle (the vehicle that is recording the data) and the agent vehicles trajectories, and the geometries of the lane boundaries. 
The first step of our data pipeline is to combine both trajectories and map information into uniform data sequences. Each data sequence is composed of multiple data frames. We further exploit the convolutional neural network (CNN) and recurrent neural network (RNN) models with self-supervised mechanisms to extract expressive frame feature vectors and then sequence feature vectors. Given these extracted features, a K-means clustering algorithm is applied for grouping the data. The evaluation of such an unsupervised learning task is non-trivial, and we propose novel techniques for this evaluation purpose. We compare our results with the most recent research study as the benchmark method and further provide our ablation study results.  

Our clustering framework has the following technical advantages. 
\begin{itemize}
    \item Our method comprehensively consider the general real-world traffic elements, including both agents and map information;
    \item We utilize a self-supervised deep learning technique for feature extraction to avoid the bias introduced by handcrafted features;
    \item A data-augmentation based external evaluation scheme is designed and exploited for analyzing our clustering results while avoiding the requirement of ground-truth labels.
\end{itemize}

\section{Related Work}

The clustering of vehicle driving data has attracted many research interests because of its importance for developing autonomous driving technologies.
Identifying the vehicle driving scenario type helps to generate representative simulation test cases.
\cite{calo2020generating} proposes a framework for generating collision avoidance scenarios in the simulation system.
\cite{nitsche2017pre} investigates the algorithms of clustering vehicle collision data based on pre-defined scenario types. 
Moreover, 
\cite{kruber2018unsupervised} discusses unsupervised learning algorithms based on the random forest for grouping the general traffic data.

Existing methods of obtaining the scenario types from vehicle driving data usually take advantage of handcrafted features, resulting in overlooking certain cases or being inadequate. Hence \cite{hauer2020clustering} proposes a method of extracting scenario types from driving data while using the human mental model as little as possible. The method in \cite{hauer2020clustering} extracts speeds and distances and compresses such features by principal component analysis (PCA). The similarity between data sequences is achieved by dynamic time wrapping (DTW). 


Recent studies in the field of video representation learning provide us alternatives for understanding vehicle data sequences. For example, \cite{srivastava2015unsupervised} proposes a model of recurrent neural network (RNN) for learning the expressive representation of a video. 
Besides, \cite{han2019video} tries to extract meaningful representation by enforcing the prediction quality from the extracted feature. 
More recently, \cite{tokmakov2020unsupervised} learns the video representation while considering the clustering quality using the extracted feature, improving the compactness of the clustering result.
We adopt the deep learning scheme with RNN for processing various-length sequences and extensively explore the self-supervision technique to extract and learn the expressive feature vectors that can represent the data sequences. Such compact representations are used for grouping the data sequence into different scenario types. 

Our work shares a similar objective of eliminating hand-crafted feature extraction \cite{hauer2020clustering}, but leveraging the unsupervised deep learning techniques for more effective spatial and temporal feature extraction. We adopt the video feature extraction techniques from \cite{han2019video,tokmakov2020unsupervised} with the improvement of enforcing the similarity property into the extracted features.
Moreover, our approach can be scaled to large-scale data set compared with the existing traffic data clustering algorithms.

\section{Approach}


\subsection{Problem Formulation}

In this article, we aim to group a large vehicle driving data-set.
Given a set of unlabelled vehicle driving data sequences 
\[
X = \{x_1, x_2, ...x_n \},
\]
where $n$ is the size of the data-set. Our objective is to design a mapping function $\Psi$ that maps a data sequence into a compact representation space $Y$, i.e.
\[
y_i = \Psi(x_i),
\]
where ${y_i} \in {\mathbb{R}}^{d}$,
so that similar data sequences can be congregated while dissimilar ones are dispersed in the $Y$ space. 
Subsequently, a clustering algorithm groups the data-set based on the extracted features $y_i$ and generates the clustering labels as
\[
l_i = \Phi(y_i),
\]
where $\Phi$ represents the clustering algorithm.

There are three major challenges for this task. The first is that the data sequence is not uniform, in the sense of each data sequence's length and the information in each frame of a data sequence. On the other hand, current clustering algorithms normally compare the difference between uniformed data representations.
The second challenge is the extraction of data features and the definition of similarity without ground-truth labeling. 
Finally, without the ground-truth labeling, the evaluation of clustering results becomes a non-trivial job. 
Towards this end, we design a framework that first unifies each frame of the raw data sequence $x_i$, so that the mathematical representation of each data frame shares the same dimension and size.
Such uniformed data representations are further compressed to conclude more compact latent space representations. Later, these uniform and compact representations are grouped by a clustering algorithm.
The evaluation of the clustering results takes advantage of data augmentation. The overall framework is shown in Fig. \ref{fig:framework}.

\begin{figure}[ht!]
    \centering
    \includegraphics[width=85mm]{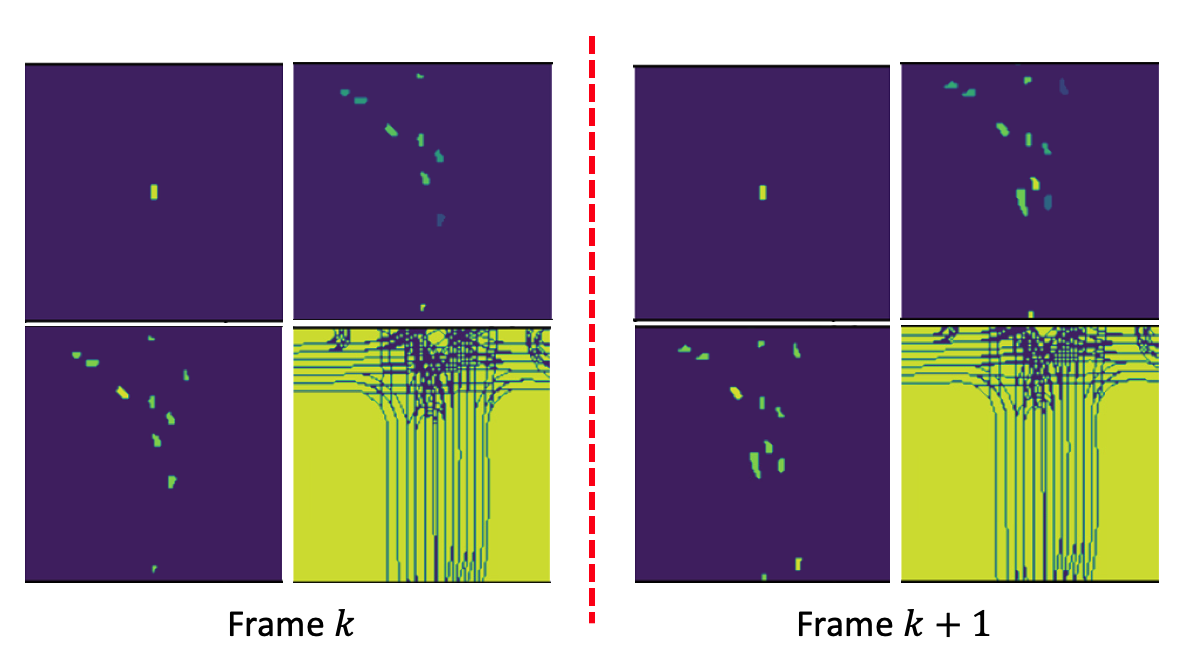} %
    \caption{Here shows the image representation for two consecutive frames. According to the Frenet coordinate, the position of the ego vehicle in the first layers remain the same, while only the color value of the pixels vary frame by frame, indicating the change of ego vehicle speed. The numbers and locations of the agent vehicles differ frame by frame, as shown in layers 2 and 3. Layer 4 displays the lane boundaries information in the surrounding environment. The orientation of the lanes rotates to cope with the Frenet coordinate system. 
    }
    \label{fig:iamge_repre}
\end{figure}

\subsection{Uniform Data Representation}

A data sequence $x_i$ can be represented by
\[
x_i = [x_i(t_{1}),~x_i(t_{2}),~x_i(t_3),...x_i(t_{m^i})],
\]
and $m^i$ indicates the size of data sequence $x_i$, where $m^i$ is not necessarily equal to $m^j$ if $i \neq j$.
Each frame of raw data sequence $x_i(t_k)$ contains the ego vehicle and agent vehicles' pose and velocity information. Meanwhile, the map information is not originally in the form of time series. We choose to render all these elements into a four-layer image to combine the vehicle poses and map information. 

Similar data representation is introduced in
\cite{henaff2019model,cui2019multimodal,Ogale-RSS-19}, but we further encode the speed information into the images 
without manually concatenation
and adopt the Frenet coordinate system \cite{werling2010optimal} for better understanding the surrounding scenarios of the ego vehicle.
The Frenet coordinate system \cite{werling2010optimal} is used in this work where the ego vehicle pose is the origin of the coordinate system, and other agent vehicles and maps are placed accordingly. Additionally, the origin of the coordinate system is placed at the center of the images. Each layer of image contains $p_m \times p_n$ pixels to represent an area of $height \times width$, hence the resolution of each pixel is $\frac{height}{p_m} \times \frac{width}{p_n}$.

The first layer contains the speed information of the ego vehicle. Given the Frenet frame, the pixels that represent the ego vehicle always located at the center of the image. The value of these pixels is the speed of the ego vehicle after normalization.

The second and third layers of the images contain the agent vehicles' relative positions and velocities. 
Pixels of each agent vehicle are obtained by projecting the agent vehicles' polygon onto the image in the Frenet coordinate. The velocity of each vehicle is decomposed w.r.t Frenet frame as well as lateral and longitudinal speeds. 
The pixels of an agent vehicle in the second image layer represents the longitudinal vehicle speed while those pixels of the third image layer indicates the lateral speeds.

The fourth image layer contains the rendering of the map's lane boundaries, which are also projected based on the Frenet coordinate.
Examples of image representation for single-frame data are shown in Fig. \ref{fig:iamge_repre}. 

\begin{figure}[bt!]
    \centering
    \includegraphics[width=60mm]{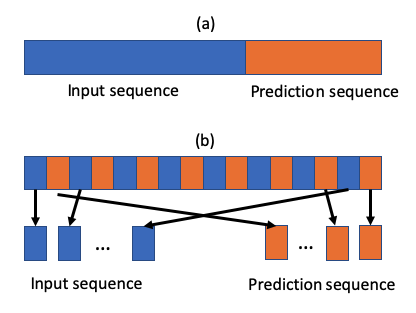}
    \caption{(a) shows a common process of preparing input and supervision data pairs for training prediction model; (b) shows the way we construct input and supervision data pairs for model training. }
    \label{fig:pre_seq}
\end{figure}

Such image representation of each frame of data is uniform, and the size of the image remains identical regardless of various numbers of agents vehicles, geometry changes of lane boundaries. 
Thus, each frame of the original data $x_i(t_k)$ is now represented by $\tilde{x}_i(t_k)$, i.e.,
\begin{align*}
    \tilde{x}_i(t_k) = R(x_i(t_k)),
\end{align*}
where $R$ represents the process of rendering.
Applying a clustering algorithm on the obtained data representation is feasible but would cost a considerable amount of computation resources. Thus further compression or feature extraction is needed.

\subsection{Spatial Data Compression}
Here, we explore deep learning techniques to 
translate the uniform frame data representation into a compact frame feature vector.
Each image representation $\tilde{x}_i(t_k)$ can be seen as a four-dimensional tensor. 
Classical dimension reduction methods such as 
principal component analysis (PCA) \cite{wold1987principal} cannot be applied for our purpose. Because PCA requires to collect all the data representation of each frame of each data sequence of the whole data-set, which is impossible in terms of memory storage and computation capability.

Alternatively, we choose a variational autoencoder (VAE) deep learning model \cite{pu2016variational} for obtaining a compact frame feature vector of each frame image, i.e.,
\begin{align*}
    z_i(t_k) = enc_c(\tilde{x}_i(t_k))\\
    \bar{x}_i(t_k) = dec_c(z_i(t_k))
\end{align*}
where $enc$ and $dec$ are encoder and decoder deep neural network, mainly composed of convolutional neural network (CNN) and fully connected layers. The specifications of the model are as follows,

When compressing such image representation into compact feature, we expect the extracted feature is expressive enough for reconstruction of the original image. Thus a reconstruction loss is applied to compare the input and output of the VAE model and we choose to minimize the mean-squared error as the reconstruction loss, such as
\begin{align*}
    \mathcal{L}_{recon} = E(\|\tilde{x}_i(t_k) - \bar{x}_i(t_k) \|^2_2).
\end{align*}

In addition to the reconstruction target, the distance between the extracted features should indicate the similarities between them. The reconstruction purpose of VAE does not guarantee the extracted feature $z_i(t_k)$ to have such property. Meanwhile, without the ground-truth labels, such property is difficult to enforce on the features. However, the availability of many data sequences provides us a self-supervised solution for the lack of labels problem. 
We introduce the contrast triplet loss \cite{dong2018triplet} between different frames during the training process, which is defined as
\begin{align*}
    \mathcal{L}^{t}_i = &E(\|z_i(t_k) - z_i(t_m) \|^2_2 - \|z_i(t_k) - z_i(t_n)\|^2_2 + \alpha ) + \\
    &E(\|z_i(t_k) - z_i(t_m) \|^2_2 - \|z_i(t_k) - z_j(t_m)\|^2_2 + \alpha ),
\end{align*}
where $\|t_k - t_m \| < \|t_k - t_n \|$ and $i \neq j$.
Intuitively, we design the triplet loss so that in the same data sequence, the closer the two frames in terms of time index $t$, the smaller the $L2$ norm of the difference between the two features. Besides, given the same difference in terms of frame index $t$, the difference between two frames inside the same sequence is smaller than that of two frames across two data sequences. 


The overall loss function is the combination of reconstruction loss and the contrast triplet loss, i.e.
\begin{align*}
    \mathcal{L} = \mathcal{L}_{recon} + \omega \mathcal{L}^t_i,
\end{align*}
where $\omega$ is a hyper-parameter of the loss ratio.
In this way, we aim to obtain an extracted feature that is both expressive for reconstruction and contains similar information between frames.

Considering each frame, the extracted is now uniform and compact, yet each data sequence's length could be different, which will be handled by the temporal model in the following section.


\subsection{Temporal Data Compression and Unsupervised Clustering}

To further compress the sequence feature representation, we exploit a recurrent neural network (RNN) based auto-encoder (AE) model.
Such process can be compactly represented as 
\begin{align*}
    y_i = enc_s(z_i)\\
    \tilde{z}_i = dec_s(y_i)\\
    \hat{z}_i = pre_s(y_i)
\end{align*}
where $z_i = [z_i(t_0), z_i(t_1), ... z_i(t_{m_i})]$, $enc_s$ indicates a RNN model for encoding purpose while $dec_s$ and $pre_s$ are RNN model for decoding purpose. As discussed in \cite{srivastava2015unsupervised}, the extracted feature of a data sequence is considered expressive when it can both reconstruct the input data sequence as well as predict the future states. Besides, \cite{srivastava2015unsupervised} has shown that reversing the order of reconstruction target improves the quality of extracted feature.

\begin{figure}[t!]
    \centering
    \includegraphics[width=80mm]{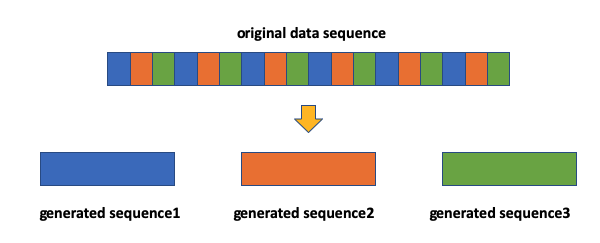}
    \caption{Augmentation of a data sequence into multiple same cluster data sequence}
    \label{fig:data_aug}
\end{figure}

Following this idea, we implement two decoder models. One decoder is designed for reconstruction, and one model is for prediction yet with modification of the prediction supervision. 
As shown in Fig. \ref{fig:pre_seq}(a), given a data sequence, the input sequence and prediction ground-truth can be obtained by slicing the original data sequence into two parts. The previous part is taken as the input data $z_i$ while the latter part is considered the prediction ground-truth $\bar{z}_i$, i.e., supervision of $\hat{z}_i$. Such an approach is introduced in \cite{srivastava2015unsupervised}.
However, such method would cause the loss of expression of the prediction part in the extracted feature $y_i$, since none of the information inside prediction sequence is processed by the encoding model. Hence, we explore an alternative that the input data sequence is composed by certain number of extracted frames from the original data sequence, while the frames next to the selected input data frames are selected to construct the prediction data sequence, as shown in Fig. \ref{fig:pre_seq}(b).

Hence, the objective is to minimize the difference between reconstruction and prediction with respect to ground-truth, such as
\begin{align*}
    \mathcal{L}^{rp}_i = E(\|z_i - \tilde{z}_i \|^2_2) + E(\|\bar{z}_i - \hat{z}_i \|^2_2).
\end{align*}

In addition to enforcing the extracted feature $y_i$ to be expressive, another desired property to enforce is that the distance between semantically similar instances should be minimized, i.e., the similar definition between data sequences. We adapt the local aggregation approach in \cite{tokmakov2020unsupervised} for this purpose. 
An iterative expectation-maximization (EM) framework is exploited. A clustering algorithm first groups the data based on the sequence feature vectors $y_i$, then a loss function is constructed based on the grouping result, and the model parameters are updated via back-propagation. The loss is defined as 
\begin{align*}
    \mathcal{L}_{i}^{la} = -\log \frac{P(C_i \cup B_i |y_i )}{P(B_i | y_i)},
\end{align*}
where $B_i$ indicates the group of $k$ data vectors closest to $y_i$, measured by Euclidean distance; $C_i$ represents the set of features grouped in the same cluster by $\Psi$. Additionally, the probability of data $y$ inside a set $A$ is defined as
\[
P(A|y) = \sum_{i \in A}P(i | y),~~~\textup{and}~~~ P(i | y) = \frac{\exp(y_i^T y /\tau)}{ \sum^N_{j=1}\exp(y_j^T y /\tau)},
\]
where $\tau$ is a hyper-parameter.
Intuitively, such loss function penalizes the situation that data features are not grouped in the same cluster but are close to each other in the sense of Euclidean distance. 


\subsection{Clustering Methods}

Here we select the K-means clustering algorithm for grouping the sequence feature vectors into clustering. 
The Silhouette coefficient is calculated and considered as an indication for the optimal number of clusters \cite{zhou2014automatic}.

\section{Evaluation Metrics}

Evaluating the clustering result of un-labeled data set is a non-trivial task. 
Many metrics, such as Silhouette coefficients and Calinski-Harabasz index, assess the clustering performance w.r.t. the extracted features, i.e., the $Y$ space in our case. However, such metrics could not reflect the quality of feature extraction $\Psi$. Thus the overall clustering quality remains undetermined. For a comprehensive evaluation of the data sequence clustering result, we exploit a data augmentation mechanism and related data sequence mining results for assessing the overall performance.

Towards this end, we prepare an evaluation data-set from the rendered data sequences $\tilde{x}_i$ by data augmentation. Given one data sequence, we sample the data frames multiple times following a uniform distribution to form multiple data sequences, as shown in Fig. \ref{fig:data_aug}. The generated data sequences are called derived data sequence. The derived sequences from the same original sequence do not contain any identical data frame, but they should belong to the same class. We call the derived data sequences from the same original data sequence as siblings.

\textbf{True Positive Rate:} Our clustering pipeline is applied to these derived data sequences to obtain the grouping results. 
A metric is defined as true positive rate, which is the ratio between the number of corrected grouped data sequences and the total number of data sequences, i.e.,
\begin{align*}
    TP = \frac{P_c}{N_d}
\end{align*}
where $N_d$ is the total number of clustered derived data sequences, and $P_c$ is the number of derived data sequences that all of its derived siblings are clustered into the same group by the algorithm.

\textbf{False Positive Rate:} In addition to the data augmentation evaluation metrics, we also take advantage of the rule-based data mining 
results of the data sequences, which are called gradings. 
The gradings are obtained by checking if the ego vehicle trajectories or the agent vehicles trajectories satisfy the pre-defined rules. 
We select several data mining results to form the grading vector, i.e.,  
\[v_i = [v_i^1, v_i^2, v_i^3, v_i^4, v_i^5, v_i^6]^T,\]
where the features are
\begin{itemize}
    \item $v_i^1$: If the ego vehicle follows an agent vehicle with favorable speed and distance;
    \item $v_i^2$: If the ego vehicle follows an agent vehicle in a junction area with favorable distance;
    \item $v_i^3$: If the ego vehicle has a risk of collision;
    \item $v_i^4$: If the jerk of the vehicle exceeds a limit;
    \item $v_i^5$: If the ego vehicle follows an agent vehicle in low speed;
    \item $v_i^6$: If the ego vehicle runs a yellow light.
\end{itemize}
From our observation, the distribution of the grading items could be quite unbalanced, for example, around $90 \%$ of the data sequences have the identical grading item $v^1_i$. Hence, the grading vectors 
could not be a meaningful representation of the data sequence but
could serve as an indicator of wrongly grouped data sequence. Namely, the clustering result is incorrect if the data sequences in the same cluster process the different grading vectors.
Therefore,
a metric named false positive rate is defined accordingly as the ratio between the number of incorrectly clustered data sequences and the total number of data sequences, i.e.,
\begin{align*}
    FP = \frac{P_w}{N_d}
\end{align*}
where $P_w$ is the number of derived data sequences whose data mining results $v$ are different from that of the other majority of the data sequences in the same cluster.

\section{Experiments}

\subsection{Dataset}


We train the frame feature extraction model and sequence feature compression model using a private data-set, recorded in various road conditions, and evaluate the results with the augmented data-set via the proposed data augmentation technique.

\textbf{Private Data-set} Our data is composed of $10,000$ data sequences collected in real-world driving scenarios by a manually driven data recording vehicle.
Each data sequence contains the ego vehicle's trajectory and the detected agent vehicles' trajectories on the road. 
Besides, a high-definition (HD) map containing information of road boundaries and lane boundaries is accessible. 
Moreover, each sequence has a corresponding rule-based grading result by expert-defined criteria, including vehicle driving safety and comfortableness related metrics as described in the previous section.

\textbf{Augmented Data-set} The evaluation metrics we proposed are based on the data augmentation mechanism. We select 1,000 data sequences as the base data-set, then augment each data sequence into $5$ derived data sequences. Hence, we form an augmented data-set of size $5000$.
To keep the similarity with source data sequences, we use a uniform sampling for data augmentation.

\textbf{Public Data-set}
In addition to the private data-set, the proposed method has been applied to the public nuScenes data-set \cite{nuscenes2019}. NuScenes contain 1000 sequences which split into 700 for training, 150 for validation and 150 for testing, respectively. Only the training and validation data are used in this paper. Each sequence lasts about 20 seconds with 20 FPS frequency and annotation at 2 FPS. NuScenes data-set provides convenient APIs to query vehicle locations, obstacle vehicle locations and map renderings. The clustering pipeline remains the same as we applied on the private data-set, including the data representation, spatial and temporal data compression. The only difference lies in the evaluation part, when the false positive rate is calculated. Since nuScenes data-set does not provide the same grading metrics, we hence need to select the features from the scenario description. 
The features we use for False Positive Rate for public-dataset nuScenes are ``wait at intersection'', ``turn right'', ``turn left'', ``cross intersection'', ``arrive at intersection'', ``parked cars'', ``parking lot'', ``nature''.

\subsection{Implementation Details}

For the uniform data representation rendering, we choose the rendering image pixel numbers as
$p_m = p_n = 129$ and the range of the representation as $height = width = 100 m$.
We select the following hyper-parameters for the training process of the frame feature extraction (CNN-VAE) model. The batch size is selected as $1600$ with $8$ Nvidia $M40$ GPUs, the learning rate is $0.001$ using Adam optimizer, the ratio $\omega$ between contrast triplet loss $\mathcal{L}^t$ and reconstruction loss $\mathcal{L}_{recon}$ is $1.0$, and the triplet loss margin $\alpha$ is chosen as $1.0$. the hyper-parameter $\tau$ is chosen as $0.07$.



For a fair comparison, we use the same training strategies and hyperparameters in all the experiments. 

We train our sequence feature compression model with Adam optimizer using a learning rate value of 0.001, and the weight decay is 0.01. The batch size is 256, and the total epoch number is 200. 
For local aggregation iterative training, we select the closest $100$ data to form the background neighbor set $B_i$. To construct the close neighbor set $C_i$, we repeatedly run the k-means cluster algorithm $4$ times to group the data into $50$, $100$, $150$, and $200$ clusters, respectively. The local aggregation loss ratio w.r.t. the reconstruction loss is $0.0001$.

To reduce the heavy burdens of training the whole model from scratch and help the training converge fast, our training is split into three stages. First of all, the frame features extractor (ConvNet) module is trained alone, then we freeze the weights of ConvNet module and training the sequence feature extractor(seq2seq) module. In the end, we fine-tune on the whole model.

\begin{table}[ht!]
 	\centering
 	\resizebox{0.46\textwidth}{!}
 	{%
 		\begin{tabular}{lccccc}
 			\hline
 			\textbf{Methods} & True Positive Rate &  False Positive Rate \\ \hline
 			Proposed Method  &  \textbf{99.88}\%  & \textbf{0.10}\%  \\ \hline
            Proposed Method \\ w/o prediction &  98.48\% & 1.92\%  \\ \hline
            Proposed Method \\ w/o reconstruction & 98.94\% & 1.40\%   \\ \hline
            Proposed Method \\ w/o triplet loss &  99.86\% & 0.30\%   \\ \hline
            Proposed Method \\ w/o Reverse-order & 97.66\% & 3.86\% \\ \hline
            Sequence Average & 95.41\% & 2.50\%  \\ \hline
            Benchmark &  77.27 \% & 3.00\%  \\ \hline
 		\end{tabular}%
 	}
 	\caption{Evaluation results on private Data-set.}
 	\label{tab:exp_private}
 \end{table}

\begin{table}[ht!]
 	\centering
 	\resizebox{0.46\textwidth}{!}
 	{%
 		\begin{tabular}{lccccc}
 			\hline
 			\textbf{Methods} & True Positive Rate &  False Positive Rate \\ \hline
 			Proposed Method  &  \textbf{98.70}\%  & \textbf{2.25}\%  \\ \hline
            Proposed Method \\ w/o prediction &   98.23\% &  2.63\%  \\ \hline
            Proposed Method \\ w/o reconstruction &  98.51\% &  2.58\%   \\ \hline
            Proposed Method \\ w/o triplet loss &   92.6\% & 5.95\%   \\ \hline
            Proposed Method \\ w/o Reverse-order &  98.63\% &  3.21\% \\ \hline
            Sequence Average &  98.58\% &  2.50\%  \\ \hline
            Benchmark &    67.10\% &  31.57\%  \\ \hline
 		\end{tabular}%
 	}
 	\caption{Evaluation results on nuScenes Data-set.}
 	\label{tab:exp_nuScenes}
 \end{table}

\subsection{Performance Evaluation}

In this section, we evaluate our proposed method by the proposed metrics (True Positive Rate and False Positive Rate), with the augmented data-set. The results are shown in Table \ref{tab:exp_private}, the True Positive Rate is 99.88\%, and the False Positive Rate is 0.10\%. We compare our performance with a benchmark method proposed in \cite{hauer2020clustering}, one of the most recent results of vehicle driving data clustering for scenario understanding. 
The benchmark method achieves the True Positive Rate of $77.27 \%$ and the False Positive Rate of $3.0 \%$. Meanwhile, the benchmark method is applied on a truncated data-set of size $100$ because the computation and storage requirements of the related PCA and DTW techniques limit the capability of using a larger data-set with this benchmark method. Our method, instead, can be applied to much larger data-sets. The evaluation result of the nuScenes data-set is shown in Table \ref{tab:exp_nuScenes}. With our proposed pipeline, the True Positive Rate is $98.70 \%$ and the False Positive Rate is $2.25\%$.

\begin{figure}[h!]
    \centering
    \includegraphics[width=85mm]{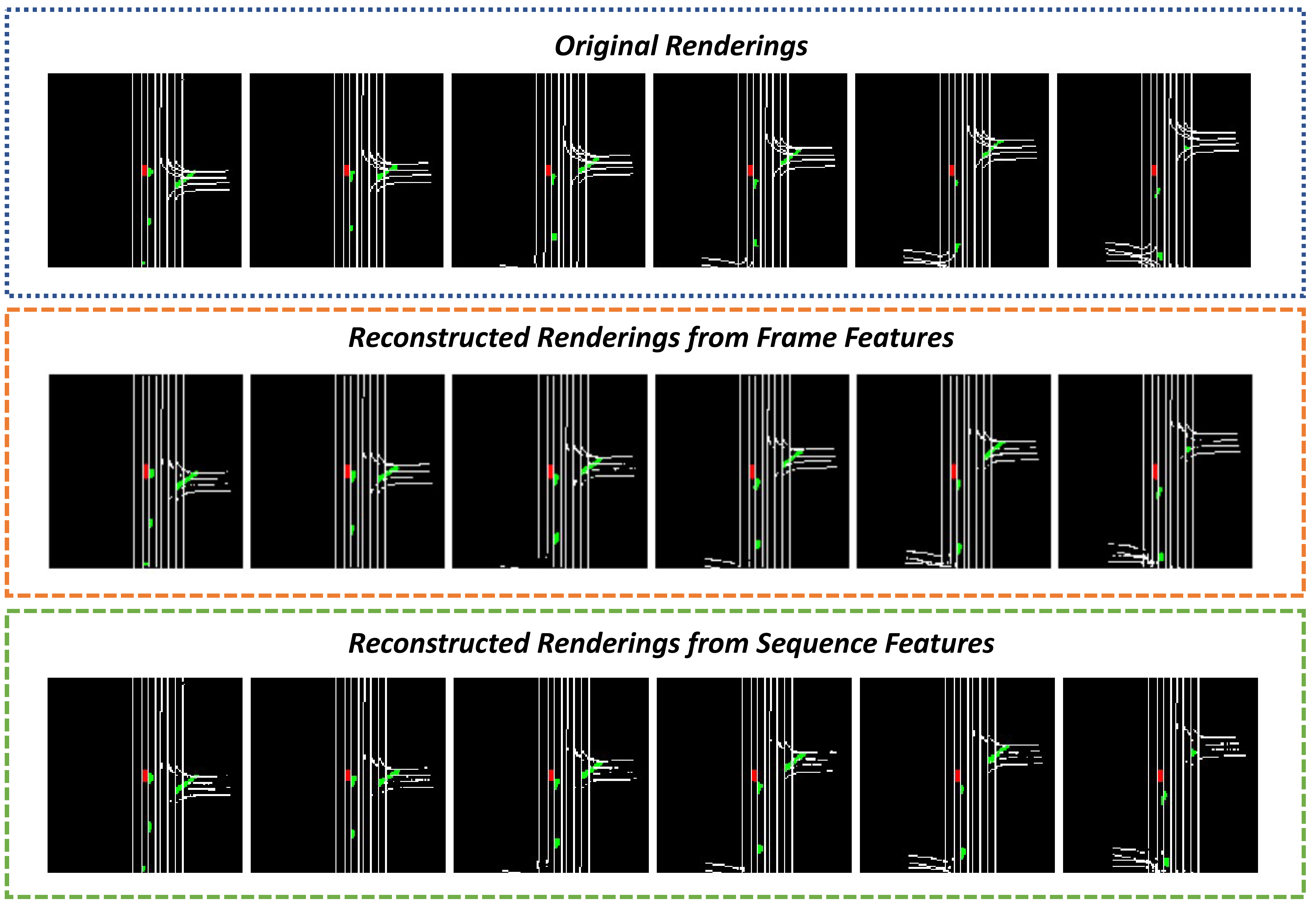}
    \caption{The first row shows the original renderings of the vehicle driving data. The second row shows the reconstructed renderings from the frame feature vectors; each image is decoded from an individual feature vector.
    The third row shows the reconstructed renderings from the sequence feature vector; that is, a single feature vector is first decoded into multiple frame feature vectors. Moreover, each frame feature vector is further reconstructed into renderings.}
    \label{fig:reconst_frame}
\end{figure}

\subsection{Analysis}

\begin{figure}[bh!]
    \centering
    \includegraphics[width=70mm]{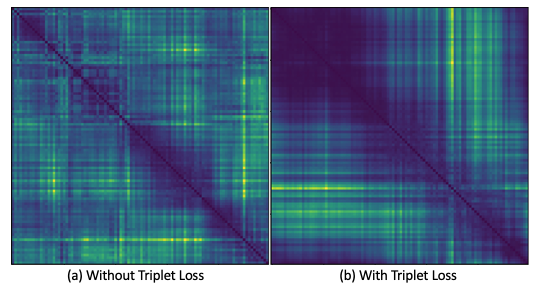}
    \caption{The effect of triplet loss: These figures show the visualization of the similarity matrices between each frame feature vectors of a data sequence. The $(i, j)$th pixel of the picture indicates the similarity between the $i$th frame and the $j$th frame. Dark pixel indicates the two corresponding frame features have small Euclidean distance and vise versa.
    (a) The frame features are extracted by a model while no triple loss is added during training; (b) The frame features are extracted by a model while triple loss is added during training. 
    Via the proposed triplet loss, when two frames are closer to each other in temporal order, their feature vectors are more similar.
    }
    \label{fig:triplet_loss}
\end{figure}

Moreover, to illustrate the effectiveness of the feature extraction models, samples of the original data renderings and the reconstructed renderings are shown in Fig. \ref{fig:reconst_frame}.
To shed more light on the effect of the proposed contrast triplet loss when training the frame extraction model, we compare the Euclidean distance between each frame feature vector when training with/without contrast triplet loss. The result is shown in Fig. \ref{fig:triplet_loss}. It can be seen that the triplet loss helps enforce the property that consecutive frame vectors are more similar to each other. Namely, the Euclidean distance between two consecutive frames is smaller than the ones separated by several frames.

\begin{figure}[bh!]
    \centering
    \includegraphics[width=60mm]{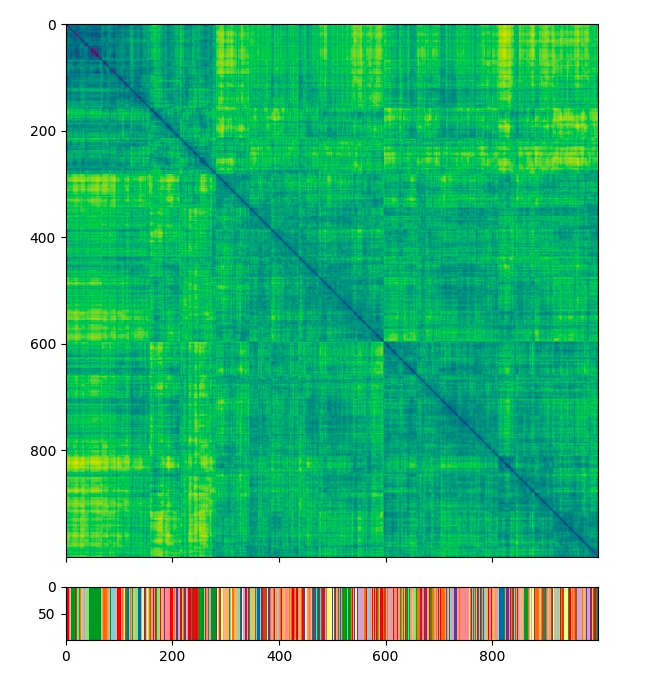}
    \caption{Clustering result of $1000$ data sequences. The upper plot shows the Euclidean distances, where the $(i, j)$th pixel indicates the distance between the $i$th data and the $j$th data. The lower color bar indicates the which cluster the $i$th data belongs to.}
    \label{fig:clustering_result}
\end{figure}

Fig. \ref{fig:clustering_result} shows clustered samples from the clustering results.  Each driving sequence is rendered as an image containing ego vehicle trajectory, agent vehicle trajectories, and map information. 
The samples from the same cluster normally comprise similar road structures such as intersection, similar vehicle behavior such as passing and yielding.


\section{Ablation Studies}
In this section, we conduct ablation studies to evaluate the necessity of each part of our architecture design. We individually remove several modules from our proposed network model. More specifically, five types of settings are evaluated here to compare with the proposed method. The experiment settings are 
{\bf (1)} Proposed method without prediction head; {\bf (2)} Proposed method without reconstruction head; {\bf (3)} Proposed method without triplet loss; {\bf (4)} Proposed method, but the order of reconstruction targets is not reversed; {\bf (5)} Averaging model, where the frame feature extractor keeps the same with the proposed method, but the sequence to sequence model is removed and replaced by a simple average pooling along the temporal dimension.

Furthermore, all the settings of these experiments strictly follow the training steps of the proposed model.
The evaluation results are shown in Table \ref{tab:exp_private} and Table \ref{tab:exp_nuScenes}. As we can see, with the private data-set, the proposed method has the best performance. The prediction and reconstruction module both play an important role in the sequence clustering. Without the prediction and reconstruction module, the True Positive Rate drop about $1.40 \%$ and $0.94 \%$, and the False Positive Rate increase about $1.82 \%$ and $1.30 \%$, respectively.
The reconstruction module helps to compress the most critical information, and the prediction module helps to learn the motion between frames.
Furthermore, reverse the target sequence order contributes about $2.22 \%$ in True Positive Rate, and the False Positive Rate decrease about $3.76 \%$, benefit from . 
The True Positive Rate of The averaging model is only $95.41 \%$, and the False Negative Rate is $2.5 \%$, which shows that the temporal information is quiet pivotal in sequence representation.
Similar results are also shown with the public NuScenes data-set, where our proposed pipeline achieves the best evaluation performance.

\section{Conclusions}

This paper proposed a method of clustering the vehicle driving data into different groups. The performance of our method vastly exceeds the existing benchmark method. Such improvement is achieved by the proposed uniform data representation, including comprehensive information on the traffic situation. Moreover, we exploit advanced deep learning techniques with self-supervised fashion for expressive feature extractions. 
In addition to the clustering framework design, we introduce an evaluation method via data augmentation to assess the performance of the un-labeled data clustering problem. 

Future developing directions are two-fold. One is to consider additional agents and map information, for example the trajectories of bikes and pedestrians, crosswalks, traffic lights, stop signs and so on. 
Towards this end, the other future work would be exploring 
more sophisticated deep learning model architectures, especially for processing the temporal data sequences, since current RNN models could forget the early frames as the sequence length extends.

\bibliographystyle{IEEEtran}
\bibliography{IEEEabrv,ref}

\begin{thebibliography}{10}
\providecommand{\url}[1]{#1}
\csname url@rmstyle\endcsname
\providecommand{\newblock}{\relax}
\providecommand{\bibinfo}[2]{#2}
\providecommand\BIBentrySTDinterwordspacing{\spaceskip=0pt\relax}
\providecommand\BIBentryALTinterwordstretchfactor{4}
\providecommand\BIBentryALTinterwordspacing{\spaceskip=\fontdimen2\font plus
\BIBentryALTinterwordstretchfactor\fontdimen3\font minus
  \fontdimen4\font\relax}
\providecommand\BIBforeignlanguage[2]{{%
\expandafter\ifx\csname l@#1\endcsname\relax
\typeout{** WARNING: IEEEtran.bst: No hyphenation pattern has been}%
\typeout{** loaded for the language `#1'. Using the pattern for}%
\typeout{** the default language instead.}%
\else
\language=\csname l@#1\endcsname
\fi
#2}}

\bibitem{waymolaunch}
P.~LeBeau, ``Waymo launches delivery service after raising $\$ 2.25$ billion,''
  \url{https://www.cnbc.com/2020/03/02/waymo-launches-delivery-service-after-raising-2point25-billion.html},
  Mar. March 2, 2020.

\bibitem{sun2020scalability}
P.~Sun, H.~Kretzschmar, X.~Dotiwalla, A.~Chouard, V.~Patnaik, P.~Tsui, J.~Guo,
  Y.~Zhou, Y.~Chai, B.~Caine, \emph{et~al.}, ``Scalability in perception for
  autonomous driving: Waymo open dataset,'' in \emph{Proceedings of the
  IEEE/CVF Conference on Computer Vision and Pattern Recognition}, 2020, pp.
  2446--2454.

\bibitem{li2019aads}
W.~Li, C.~Pan, R.~Zhang, J.~Ren, Y.~Ma, J.~Fang, F.~Yan, Q.~Geng, X.~Huang,
  H.~Gong, \emph{et~al.}, ``Aads: Augmented autonomous driving simulation using
  data-driven algorithms,'' \emph{Science Robotics}, vol.~4, no.~28, 2019.

\bibitem{Koopman16}
P.~Koopman and M.~Wagner, ``Challenges in autonomous vehicle testing and
  validation,'' \emph{SAE International Journal of Transportation Safety},
  vol.~4, pp. 15--24, 04 2016.

\bibitem{hauer2020clustering}
F.~Hauer, I.~Gerostathopoulos, T.~Schmidt, and A.~Pretschner, ``Clustering
  traffic scenarios using mental models as little as possible,'' in
  \emph{PrePrint for the proceedings of IEEE Intelligent Vehicles Symposium
  2020}, 2020.

\bibitem{bach2016model}
J.~Bach, S.~Otten, and E.~Sax, ``Model based scenario specification for
  development and test of automated driving functions,'' in \emph{2016 IEEE
  Intelligent Vehicles Symposium (IV)}.\hskip 1em plus 0.5em minus 0.4em\relax
  IEEE, 2016, pp. 1149--1155.

\bibitem{hauer2019did}
F.~Hauer, T.~Schmidt, B.~Holzm{\"u}ller, and A.~Pretschner, ``Did we test all
  scenarios for automated and autonomous driving systems?'' in \emph{2019 IEEE
  Intelligent Transportation Systems Conference (ITSC)}.\hskip 1em plus 0.5em
  minus 0.4em\relax IEEE, 2019, pp. 2950--2955.

\bibitem{apolloplatform}
{Baidu Apollo team}, ``Apollo: Open source autonomous driving,''
  \url{https://github.com/ApolloAuto/apollo}, 2020, accessed: 2019-02-11.

\bibitem{xu2015comprehensive}
D.~Xu and Y.~Tian, ``A comprehensive survey of clustering algorithms,''
  \emph{Annals of Data Science}, vol.~2, no.~2, pp. 165--193, 2015.

\bibitem{xu2005survey}
R.~Xu and D.~Wunsch, ``Survey of clustering algorithms,'' \emph{IEEE
  Transactions on neural networks}, vol.~16, no.~3, pp. 645--678, 2005.

\bibitem{nguyen2019feature}
T.~T. Nguyen, P.~Krishnakumari, S.~C. Calvert, H.~L. Vu, and H.~Van~Lint,
  ``Feature extraction and clustering analysis of highway congestion,''
  \emph{Transportation Research Part C: Emerging Technologies}, vol. 100, pp.
  238--258, 2019.

\bibitem{kruber2018unsupervised}
F.~Kruber, J.~Wurst, and M.~Botsch, ``An unsupervised random forest clustering
  technique for automatic traffic scenario categorization,'' in \emph{2018 21st
  International Conference on Intelligent Transportation Systems (ITSC)}.\hskip
  1em plus 0.5em minus 0.4em\relax IEEE, 2018, pp. 2811--2818.

\bibitem{everitt2011cluster}
B.~S. Everitt, S.~Landau, M.~Leese, and D.~Stahl, ``Cluster analysis 5th ed,''
  2011.

\bibitem{amigo2009comparison}
E.~Amig{\'o}, J.~Gonzalo, J.~Artiles, and F.~Verdejo, ``A comparison of
  extrinsic clustering evaluation metrics based on formal constraints,''
  \emph{Information retrieval}, vol.~12, no.~4, pp. 461--486, 2009.

\bibitem{calo2020generating}
A.~Cal{\`o}, P.~Arcaini, S.~Ali, F.~Hauer, and F.~Ishikawa, ``Generating
  avoidable collision scenarios for testing autonomous driving systems,'' in
  \emph{2020 IEEE 13th International Conference on Software Testing, Validation
  and Verification (ICST)}.\hskip 1em plus 0.5em minus 0.4em\relax IEEE, 2020,
  pp. 375--386.

\bibitem{nitsche2017pre}
P.~Nitsche, P.~Thomas, R.~Stuetz, and R.~Welsh, ``Pre-crash scenarios at road
  junctions: A clustering method for car crash data,'' \emph{Accident Analysis
  \& Prevention}, vol. 107, pp. 137--151, 2017.

\bibitem{srivastava2015unsupervised}
N.~Srivastava, E.~Mansimov, and R.~Salakhudinov, ``Unsupervised learning of
  video representations using lstms,'' in \emph{International conference on
  machine learning}, 2015, pp. 843--852.

\bibitem{han2019video}
T.~Han, W.~Xie, and A.~Zisserman, ``Video representation learning by dense
  predictive coding,'' in \emph{Proceedings of the IEEE International
  Conference on Computer Vision Workshops}, 2019, pp. 0--0.

\bibitem{tokmakov2020unsupervised}
P.~Tokmakov, M.~Hebert, and C.~Schmid, ``Unsupervised learning of video
  representations via dense trajectory clustering,'' \emph{arXiv preprint
  arXiv:2006.15731}, 2020.

\bibitem{henaff2019model}
M.~Henaff, A.~Canziani, and Y.~LeCun, ``Model-predictive policy learning with
  uncertainty regularization for driving in dense traffic,'' \emph{arXiv
  preprint arXiv:1901.02705}, 2019.

\bibitem{cui2019multimodal}
H.~Cui, V.~Radosavljevic, F.-C. Chou, T.-H. Lin, T.~Nguyen, T.-K. Huang,
  J.~Schneider, and N.~Djuric, ``Multimodal trajectory predictions for
  autonomous driving using deep convolutional networks,'' in \emph{2019
  International Conference on Robotics and Automation (ICRA)}.\hskip 1em plus
  0.5em minus 0.4em\relax IEEE, 2019, pp. 2090--2096.

\bibitem{Ogale-RSS-19}
M.~Bansal, A.~Krizhevsky, and A.~Ogale, ``Chauffeurnet: Learning to drive by
  imitating the best and synthesizing the worst,'' in \emph{Proceedings of
  Robotics: Science and Systems}, FreiburgimBreisgau, Germany, June 2019.

\bibitem{werling2010optimal}
M.~Werling, J.~Ziegler, S.~Kammel, and S.~Thrun, ``Optimal trajectory
  generation for dynamic street scenarios in a frenet frame,'' in \emph{2010
  IEEE International Conference on Robotics and Automation}.\hskip 1em plus
  0.5em minus 0.4em\relax IEEE, 2010, pp. 987--993.

\bibitem{wold1987principal}
S.~Wold, K.~Esbensen, and P.~Geladi, ``Principal component analysis,''
  \emph{Chemometrics and intelligent laboratory systems}, vol.~2, no. 1-3, pp.
  37--52, 1987.

\bibitem{pu2016variational}
Y.~Pu, Z.~Gan, R.~Henao, X.~Yuan, C.~Li, A.~Stevens, and L.~Carin,
  ``Variational autoencoder for deep learning of images, labels and captions,''
  in \emph{Advances in neural information processing systems}, 2016, pp.
  2352--2360.

\bibitem{dong2018triplet}
X.~Dong and J.~Shen, ``Triplet loss in siamese network for object tracking,''
  in \emph{Proceedings of the European Conference on Computer Vision (ECCV)},
  2018, pp. 459--474.

\bibitem{zhou2014automatic}
H.~B. Zhou and J.~T. Gao, ``Automatic method for determining cluster number
  based on silhouette coefficient,'' in \emph{Advanced Materials Research},
  vol. 951.\hskip 1em plus 0.5em minus 0.4em\relax Trans Tech Publ, 2014, pp.
  227--230.

\bibitem{nuscenes2019}
H.~Caesar, V.~Bankiti, A.~H. Lang, S.~Vora, V.~E. Liong, Q.~Xu, A.~Krishnan,
  Y.~Pan, G.~Baldan, and O.~Beijbom, ``nuscenes: A multimodal dataset for
  autonomous driving,'' \emph{arXiv preprint arXiv:1903.11027}, 2019.

\end{thebibliography}

\end{document}